\documentclass{Interspeech}

\interspeechcameraready

\title{Word stress in self-supervised speech models: A cross-linguistic comparison}

\author[affiliation={1}]{Martijn}{Bentum}
\author[affiliation={1}]{Louis}{ten Bosch}
\author[affiliation={2}]{Tomas O.}{Lentz}

\affiliation{Centre for Language Studies}{Radboud University}{Netherlands}
\affiliation{Department of Communication and Cognition}{Tilburg University}{Netherlands}
\email{martijn.bentum@ru.nl, louis.tenbosch@ru.nl, t.o.lentz@tilburguniversity.edu}

\usepackage{comment}
\usepackage{tipa}

\begin{document}

\maketitle

\begin{abstract}
In this paper we study word stress representations learned by self-supervised speech models (S3M), specifically the Wav2vec 2.0 model. We investigate the S3M representations of word stress for five different languages: Three languages with variable or lexical stress (Dutch, English and German) and two languages with fixed or demarcative stress (Hungarian and Polish). We train diagnostic stress classifiers on S3M embeddings and show that they can distinguish between stressed and unstressed syllables in read-aloud short sentences with high accuracy. We also tested language-specificity effects of S3M word stress. The results indicate that the word stress representations are language-specific, with a greater difference between the set of variable versus the set of fixed stressed languages.

\end{abstract}

\section{Introduction}
\label{introduction}

Self-supervised speech models (S3Ms) learn to represent spoken language and can be effectively fine-tuned for downstream tasks such as automatic speech recognition \cite{baevski2020wav2vec}, speaker identification \cite{chen2022large}, and emotion recognition \cite{atmaja2022evaluating}. However, their end-to-end nature makes their inner workings difficult to interpret. One approach to improve interpretability is to analyze the content of S3M layers by training a simple diagnostic classifier \cite{alain2018understanding,conneau-etal-2018-cram} to distinguish linguistically relevant units (e.g., phone labels). This method enables the investigation of speech representations across different linguistic levels, including phonetic \cite{cormacenglishDomainInformedProbingWav2vec2022}, semantic \cite{choiSelfSupervisedSpeechRepresentations2024}, syntactic \cite{shenWaveSyntaxProbing2023a}, and prosodic information \cite{bentum2024processing}. In this study, we extend this approach to examine how word stress is represented in S3Ms across multiple languages in connected speech and compare language specificity of word stress representations.

Word stress refers to the phenomenon in which certain syllables in a word are pronounced more prominently than others \cite{gussenhoven2004phonology}. The pattern of stressed and unstressed syllables provides valuable cues for a listener or speech model. In languages with fixed or demarcative stress, such as Hungarian \cite{rounds2009hungarian}, the stressed syllable occurs in a fixed position relative to the word boundary (e.g., the first syllable in Hungarian), serving as a reliable cue for word segmentation. In contrast, languages with variable or lexical stress, such as Dutch \cite{booij1999phonology}, exhibit variable stress placement with respect to word boundary, which still aids segmentation \cite{cutler1988role} but also enables minimal pairs distinguished solely by stress, such as \textipa{/'ka:nOn/} (‘cultural collection’) vs. \textipa{/ka:'nOn/} (‘artillery gun’) in Dutch. Additionally, while monosyllabic words are generally stressed, function words (e.g., articles, pronouns, particles) are typically unstressed \cite{bossong1999word}, allowing word stress to serve as a cue for part-of-speech classification.


Syllabic prominence, or stress, is phonetically realized through increased articulatory effort, resulting in syllables that are typically longer, louder (i.e., greater acoustic energy), higher in pitch, characterized by a shallower spectral tilt, and less affected by co-articulation \cite{van2018acoustic,fry1955duration,sluijter1996spectral}. However, the prominence and reliability of these acoustic correlates vary across languages \cite{gordon2017acoustic} and are not exclusively indicators of word stress. For instance, syllable duration is influenced by speech rate \cite{crystal1990articulation}, pitch by phrase accent \cite{gussenhoven2004phonology}, and co-articulation by speech style \cite{farnetani1997coarticulation}. As a result, whether a syllable is perceived as stressed also depends on contextual factors.

The salience and reliability of acoustic correlates of stress depend on language-specific variability of word stress patterns. For instance, \cite{malisz2018lexical} identified some acoustic correlates of stress in Polish, though fewer than expected compared to languages with more variable stress patterns \cite{bossong1999word}. Additionally, the predictability of stress patterns correlates with relative stress deafness in human listeners \cite{peperkamp2010perception}. Native speakers of languages with fixed stress, such as Hungarian, exhibit greater stress deafness than those of Spanish, which has more variable stress pattern. Polish, with its limited variability, shows an intermediate pattern.

Word stress provides valuable cues for speech processing that differ across languages and are salient for human listeners. In previous work  \cite{bentum2024processing}, we showed that S3Ms encode word stress representations for English words spoken in isolation. We extend this work by examining S3M word stress representations in connected speech across multiple languages. We focus on three languages with variable word stress (Dutch, German, and English \cite{booij1999phonology,wiese2000phonology,mcmahon2020introduction}) and two with fixed word stress (Hungarian and Polish \cite{rounds2009hungarian,rubach1985polish}). Based on prior research, we hypothesize that word stress representations will show greater divergence between fixed stress and variable-stress languages than among languages within each group, as differences in acoustic realization, stress-related cues, and listener sensitivity vary between these language types. 

We limit our study to bisyllabic words in short, read-aloud sentences. Due to a lack of corpora with relevant prosodic annotations, we use a lookup- and rule-based approach for automatic stress labeling. Focusing on bisyllabic words makes automatic labeling more manageable by allowing us to ignore secondary stress and distinctions between stressed and unstressed monosyllabic words. However, our labeling approach does not account for phrasal accent or stress shift (i.e., adjusting stress to avoid consecutive stressed syllables), which may introduce confounding factors. We analyze a single multilingual S3M pre-trained on the five target languages, among others. Having been exposed to diverse stress patterns during pre-training, the model may either map word stress representations to a single cross-lingual construct or maintain language-specific distinctions.

For the current study, our aim is to answer the following research questions: Can we find word stress representations in S3M embeddings based on connected speech? Can we find these stress representations in multiple languages? Do the stress representations differ between languages and more specifically between languages with variable versus fixed stress? 

\section{Methods}
\label{methods}

We used the multilingual Wav2vec 2.0 XLS-R model\footnote{\url{https://huggingface.co/facebook/wav2vec2-xls-r-300m}}\cite{babu2021xls}, henceforth denoted W2V. This model is pre-trained on 500,000 hours of speech recordings across 128 languages, including the five languages featuring
 in our study.

\subsection{Materials}
\label{materials}
We used language materials from the Common Voice corpus \cite{ardila2019common}, which consists of recordings of short read-aloud sentences. Recordings are made via public participation and a subset of recording-transcription pairs is validated for accuracy. We selected the validated sentences from the following languages: Dutch, English, German, Polish and Hungarian, see Table \ref{tab:alignment_counts}. Based on the recording-transcription pairs, we created forced-aligned phonetic transcriptions with syllable-boundary annotations with the help of the web-based MAUS forced aligner\footnote{\url{https://clarin.phonetik.uni-muenchen.de/BASWebServices/interface/WebMAUSBasic}} \cite{kisler2017multilingual}. 

We automatically generated ground-truth stress labels for each syllable in the Common Voice materials. For the languages with variable stress (Dutch, English and German), we used the lexical database CELEX \cite{baayen1996celex} to lookup the word stress status of each syllable in each word. For this, we first aligned the aforementioned phonetic transcriptions of the Common Voice materials and the CELEX transcriptions by means of the Needlemann-Wunch algorithm \cite{NEEDLEMAN1970443} and subsequently assigned the stress label to the matching syllable. The transcription alignment step is required, since the forced-aligned transcription does not necessarily match the canonical transcription in CELEX. For the languages with fixed stress (Polish and Hungarian), we used a rule-based approach\footnote{\url{https://github.com/martijnbentum/stress-miniature-succotash}}. For Polish, word stress typically falls on the penultimate
syllable, with few exceptions \cite{rubach1985polish}, and for Hungarian word stress generally falls on the first syllable \cite{rounds2009hungarian}. All Polish and Hungarian bisyllabic words in our dataset were assigned primary stress on the first syllable.

\begin{table}[th]
  \caption{Descriptive statistics of the selected bisyllabic words in the Common Voice materials for Dutch, English, German, Polish and Hungarian. Number of bisyllabic words in thousands (Word Count), number of hours of selected materials (Hours), percentage of words with stress on the first syllable (SFS) and the Common Voice version (Version).}
  \label{tab:alignment_counts}
  \centering
  \begin{tabular}{r|r|r|r|r}
    \toprule
    \multicolumn{1}{r}{\textbf{Language}} & \multicolumn{1}{c}{\textbf{Word Count}} & \multicolumn{1}{c}{\textbf{Hours}} & \multicolumn{1}{c}{\textbf{\% SFS}} & \multicolumn{1}{c}{\textbf{Version}}\\
    \midrule
    Dutch        & 158K &  20 & 75 \% & 15 \\
    English      & 523K &  96 & 82 \% & 1  \\
    German       & 265K &  66 & 84 \% & 3  \\
    Polish       & 321K &  37 & 100 \% & 15 \\
    Hungarian    & 105K &  12 & 100 \% & 15 \\
    \bottomrule
  \end{tabular}
\end{table}

\subsection{Feature extraction}
\label{feature_extraction}
For all analyses, we focus exclusively on the vowel of each syllable and exclude all words with diphthong vowels. We computed several acoustic features: \emph{Duration} was based on the forced aligned transcription of the Common Voice materials; \emph{Intensity} in dB was computed as follows \(10 \log_{10}(\overline{x^2} / 4 * 10^{-10})\), whereby \(x\) denotes the audio samples corresponding to the vowel; mean \emph{Pitch} for each vowel was computed with the Librosa Python package \cite{mcfee2015librosa} (version 0.10.2); spectral-tilt was defined according to \cite{sluijter1996spectral} as the intensity of the following four frequency bands (0 - 500, 500 - 1000, 1000 - 2000, 2000 - 4000); The peripherality of \emph{formants} was implemented as the Euclidean distance of the mean F1 and mean F2 compared to the overall mean F1 and mean F2 for a given language; and \emph{combined-features} was created by concatenating all aforementioned features into a single vector. 

Furthermore, we computed W2V model specific features in the following manner. We applied the W2V model to each sentence recording in the materials and stored the relevant frames for the CNN output, as well as the evenly sampled
transformer layers 5, 11, 17 and 23. Furthermore, we mapped the CNN output to the corresponding codevector (a discrete representation used during pre-training). The W2V model segments the audio input in frames with a duration of 25 ms and 
a strides of 20 ms. To create vowel-specific embeddings for each model layer, we extracted frames with a minimal overlap of 50\% with a vowel, based on the forced-aligned phonetic transcriptions, and applied mean pooling.

\subsection{Stress classification}
\label{stress_classification}
For all classifiers described below, we used 20-fold cross-validation with a 2/3 train, 1/3 test split. Each classifier is trained on a single feature (e.g. duration) and materials from one specific language. For the 1-dimensional acoustic features (i.e. intensity, duration, formants and pitch), we trained stress classifiers with a kernel density estimator (as implemented in the Scipy toolkit, version 1.13.0). For the multidimensional acoustic features (i.e. spectral-balance and combined-features), we applied linear discriminant analysis (LDA) as implemented in the Scikit-learn toolkit, version 1.4.0. For the W2V outputs, we trained multilayer Perceptron (MLP) classifiers as implemented in the Scikit-learn toolkit.



To measure the performance of the stress classifiers, we report the \emph{Matthews correlation coefficient} (MCC) \cite{matthews1975comparison,baldi2000assessing}. The MCC ranges from [-1,1], with a score of 0 for chance performance. MCC only reports high performance when a classifier achieves
both a good recall and precision \cite{chicco2020advantages}.

\subsection{Language comparisons}
\label{language_comparisons}
We tested the language-specific stress classifiers on the materials of all languages and, subsequently, we pooled the language-specific performance and compared it with the pooled cross-lingual performance. Additionally, we applied two clustering techniques to investigate language similarity with respect to stress. For this, we used the 20 stress classifiers trained for each feature - language combination in the 20 fold cross-validation. We applied each classifier to all languages, creating performance vectors for each classifier. We applied LDA on the performance vectors for the acoustic features and the different model layers. Furthermore, we employed agglomerative hierarchical clustering (as implemented in the Scipy toolkit) for the best performing acoustic feature and model layer.


\section{Results}
\label{results}

\subsection{Word stress representations in connected speech}
\label{connected_speech}

\begin{figure}[t]
  \centering
  \includegraphics[width=\linewidth]{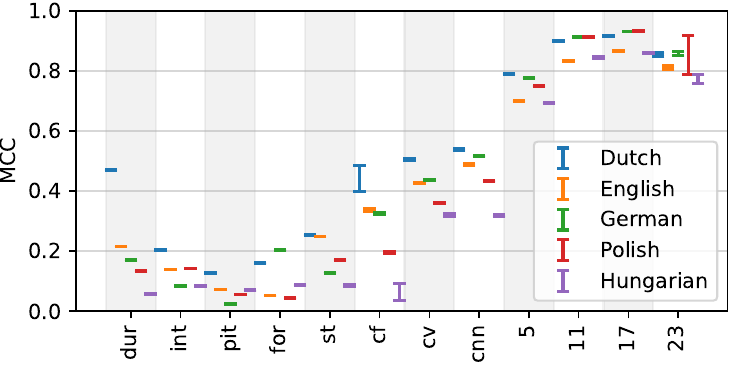}
  \caption{The performance in terms of Matthew's correlation coefficient (MCC) of classifiers trained on acoustic correlates of stress and various layers of the W2V model for Dutch, German, English, Polish and Hungarian. Results are shown for duration (dur), intensity (int), pitch (pit), formants (for), spectral-tilt (st), combined-features (cf), codevectors (cv), cnn and transformer layers 5 - 23. Error bars contain the 99\% CI of the mean.}
  \label{fig:classification_all_languages}
\end{figure}

The results for the language-specific stress classification test are summarized in Figure \ref{fig:classification_all_languages}. The error bars indicate the 99\% interval of the mean stress classification performance. For each of the five languages (Dutch, English, German, Polish and Hungarian), we found evidence for stress representations in the W2V model, with the strongest performance at transformer layer 17. The drop-off in performance at the deepest layers is well documented (e.g. \cite{pasad2021layer}) and linked to the training objective of predicting masked quantized CNN representations (i.e. codevectors), resulting in autoencoder-style behavior, whereby the deepest layers tend to become more similar to the input.

Performance with acoustic features was quite poor compared to performance with the W2V model, with combined features showing the best overall stress classification performance. For Dutch, Polish and Hungarian, the combined-features performance is similar to the best performing with a single acoustic correlate, while for German and English, combined-features shows some improvement over singleton features. Performance with most singleton acoustic features is poor, with the exception of \emph{duration} for Dutch. Performance with the acoustic features for both Polish and Hungarian is quite poor across the board, while with the best transformer layer (i.e. 17) stress classification performance was possible at a similar level as for Dutch, English and German.

\subsection{Cross-lingual word stress representations comparisons}
\label{cross-lingual}

\begin{figure}[t]
  \centering
  \includegraphics[width=\linewidth]{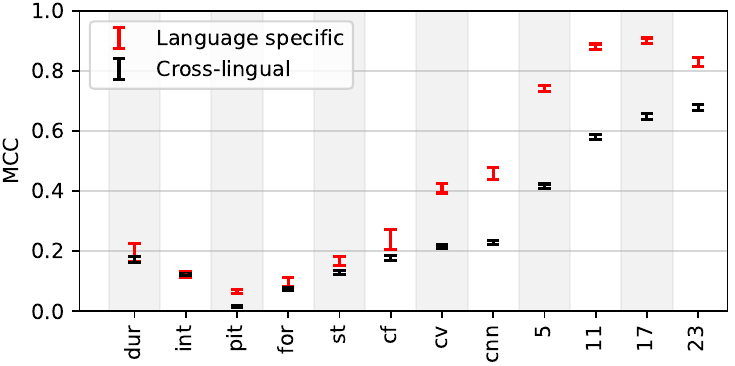}
  \caption{The pooled performance in terms of Matthew's correlation coefficient (MCC) of stress classifiers trained and tested on a specific language (red) versus cross-lingual performance (black), i.e. classifiers trained on a specific language and tested on all other languages. Results are shown for duration (dur), intensity (int), pitch (pit), formants (for), spectral-tilt (st), combined-features (cf), codevectors (cv), cnn and transformer layers 5 - 23. Error bars contain the 99\% CI of the mean.}

  \label{fig:language_specific_vs_cross-lingual}
\end{figure}

The results of the language-specificity test for stress representations in the W2V model are summarized in Figure \ref{fig:language_specific_vs_cross-lingual}. We compared the grouped performance of all language-specific stress classifiers on their target language versus the set of other languages. In the different model layers we observe language specificity, whereby the classifiers perform better if they are tested on their target-language compared to the set of other languages. Interestingly, the best performance for the cross-lingual test is for the deepest layer, which also shows less of a language-specific effect. The acoustic features only moderately show language specificity, with a small performance increase for pitch, spectral tilt, and combined features when tested on the target language versus other languages, suggesting that acoustic realization of stress do not generalize well across languages.

\subsection{Variable versus fixed stressed languages}
\label{variable_v_fixed}

\begin{figure*}[t]
  \centering
  \includegraphics[width=\textwidth]{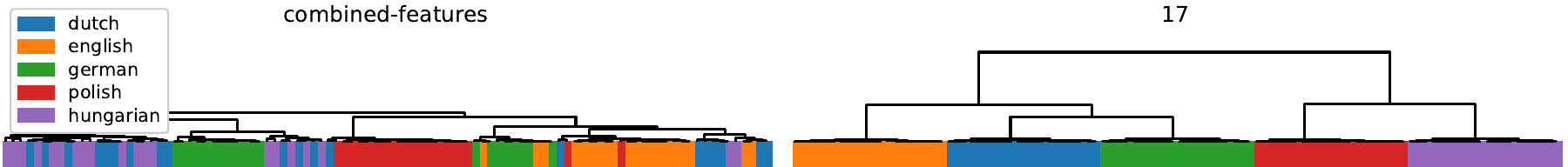}
  \caption{Resulting clusters from applying Agglomerative Hierarchical Clustering on the  classifier performance vectors based on the best performing acoustic correlate (left) and best performing transformer layer (right).}
  \label{fig:dendrogram}

\end{figure*}

\begin{figure}[t]
  \centering
  \includegraphics[width=\linewidth]{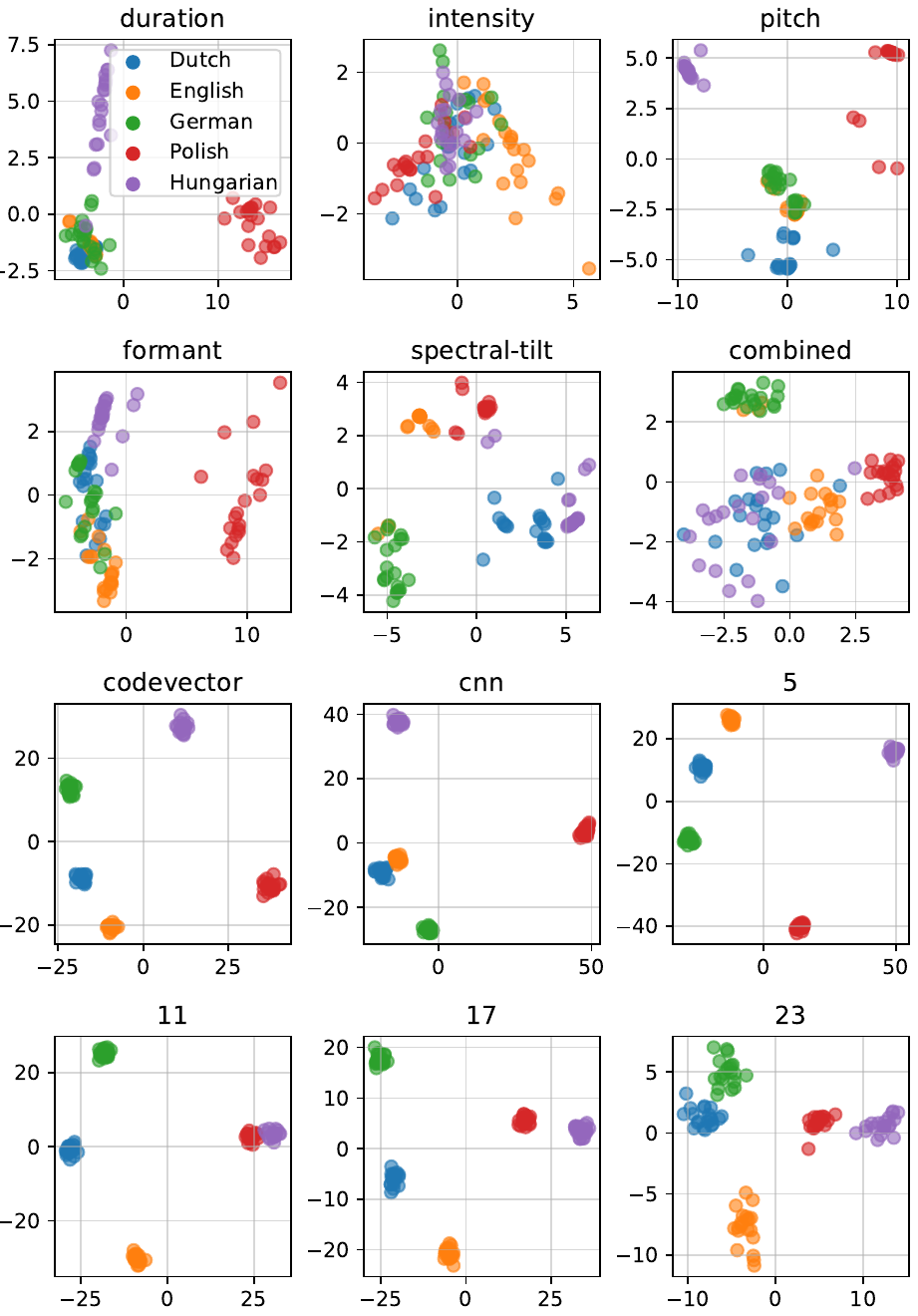}
  \caption{LDA-scatterplots based on language specific and cross-lingual performance of stress classifiers for acoustic correlates, codevectors, CNN output and transformer layers. The x- and y-axes represent the first two linear discriminants respectively. Each dot represents 1 fold of the 20-fold cross validation.}

  \label{fig:lda_clustering}
\end{figure}

The LDA results (section~\ref{stress_classification}), reflecting language-specific effects with respect to word stress classification are visualized in Figure \ref{fig:lda_clustering}. The first six panels show the acoustic feature results, with no clear separation of the different languages. To some extent, the variable stressed languages cluster more closely, for example, for \emph{duration}, Dutch, German and English are clustered closely together. Furthermore, for most acoustic features (except intensity), Polish appears to behave differently compared to the other languages. Lastly, intensity does not reveal any clear language-specific pattern. In contrast to the acoustic features, each layer of the W2V model shows a clear separation of all of the five languages, which drops off a little at the deepest layer. At the shallow layers (codevector, CNN and transformer layer 5) Dutch, German and English cluster more closely. At the deeper layers Polish and Hungarian cluster more closely.

In addition to LDA, we utilized agglomerative hierarchical clustering to test whether variable and fixed stressed languages will cluster together more within group compared to across groups. The result of the clustering for combined-features and transformer layer 17 is visualized in a dendrogram in Fig.~\ref{fig:dendrogram}. For the combined-features, languages do not cluster. There are some larger blocks for Polish, German and English, but for each there is some mixing with other languages. For transformer layer 17 there is language based clustering. Each language is grouped on a single master branch. In addition, clustering reveals that in the first split, variable stressed languages are grouped together versus the group of fixed stress languages.

\section{Discussion}

We examined word stress representations in a self-supervised speech model (S3M) across five languages. Our study focused on three languages with variable (i.e. lexical) word stress (Dutch, English, and German) and two with fixed (i.e. demarcative) word stress (Polish and Hungarian). Using short, read-aloud sentences, we applied the model at the sentence level. We selected bisyllabic words and extracted representations from different model layers corresponding to the vowels of stressed and unstressed syllables. These representations served as input for training and testing MLP-based word stress classifiers. Additionally, we extracted acoustic features known to correlate with word stress and trained separate classifiers using these features. Our results indicate that word stress is represented in the S3M for all five languages (Section \ref{connected_speech}).

The acoustic feature results (Fig.~\ref{fig:classification_all_languages}) confirm that the importance of different acoustic correlates of stress vary across languages \cite{gordon2017acoustic}. For instance, \emph{duration} is a key predictor for Dutch, while \emph{spectral-tilt} plays an important role in both Dutch and English. Additionally, consistent with \cite{malisz2018lexical}, we found that in fixed stress languages, acoustic features are less reliable for predicting word stress, with Hungarian showing the weakest predictive power and Polish falling in between. Finally, the similar classification performance of acoustic features and CNN outputs suggests that this model layer primarily captures phonetic information, aligning with findings from \cite{dieck22_interspeech,abdullah2023information,ten2023phonemic}.

The results from the language comparison experiments (Section \ref{cross-lingual}) suggest some language specificity at the acoustic feature level. However, stress classifier performance on the target language was generally comparable to that on other languages (Fig.~\ref{fig:language_specific_vs_cross-lingual}). This might also be related to the poor overall predictive performance for word stress of the acoustic features. In contrast, the model layer results show a strong language-specific effect, with a substantial drop in performance when tested on the set of other languages. Notably, this difference decreases in deeper layers, suggesting that word stress representations become more abstract, consistent with findings from \cite{bentum2024processing}.

The clustering results  (see Figure \ref{fig:lda_clustering}) show that languages in our study
can be clearly separated at all model layers. This contrasts with \cite{dieck22_interspeech}, who found no language specificity at the CNN layer for phone representations but did observe it in higher transformer layers. This discrepancy may stem from differences in the linguistic feature analyzed (phones vs.\ stress) or from the fact that our findings are based on classification performance, which aggregates results across many items (i.e., vowel tokens), rather than directly analyzing representations. Our results suggest that word stress representations remain language-specific at every model layer, whereas acoustic features do not show the same pattern. This could be partly attributed to the transformer’s large receptive field, which allows it to incorporate broader contextual information, while our acoustic features are derived solely from the vowel segment. However, this is not the sole explanation, as both the codevector and CNN, despite their narrower receptive fields, already exhibit strong language separation.

Our findings support the hypothesis that word stress representations differ between languages with variable stress and those with fixed stress (Section \ref{variable_v_fixed}). The W2V model used in our experiments is pre-trained on 128 languages, exposing it to a wide range of language-specific stress patterns, or their absence. While one might expect the model to have less reliable word stress representations for fixed stress languages, because of less reliable acoustic cues \cite{malisz2018lexical}, potentially mirroring the stress deafness observed in native human listeners \cite{peperkamp2010perception}, this appears not to be the case. Instead, word stress can still be predicted with high accuracy based on model representations, even for fixed stress languages. However, to the extent the W2V model encodes word boundaries independently of word stress, such as through phonotactic cues, it could serve as a highly reliable predictor of word stress in fixed stress languages and would be a confounding factor.

Our study provides evidence that S3Ms encode word stress representations across multiple languages and distinguish between variable and fixed stress languages. Future studies could investigate factors such as secondary stress, stress shift, or phrase accent. Additionally, the relative importance and causal order of word boundary and word stress representations should be further studied. 


\section{Acknowledgements}

All authors participate in the Dutch NWO/NWA project InDeep (https://www.nwo.nl/en/projects/nwa129219399), led by J. Zuidema (Univ. of Amsterdam).

\bibliographystyle{IEEEtran}
\bibliography{main}

\end{document}